\documentclass[]{x2lab}
\usepackage{pifont}  
\usepackage{microtype}
\usepackage{graphicx}
\usepackage{subcaption}
\usepackage{csquotes}
\usepackage{afterpage}
\usepackage[framemethod=TikZ]{mdframed} 
\usepackage{amsmath}
\usepackage{amssymb}
\usepackage{mathtools}
\usepackage{amsthm}
\usepackage{xspace}
\usepackage{colortbl}
\usepackage{multirow}
\usepackage{enumitem}
\usepackage{wrapfig}
\usepackage{nicefrac}
\usepackage{makecell}
\usepackage{multirow}
\usepackage{xcolor}
\usepackage{mdframed}

\usepackage{pifont}
\definecolor{paperblue}{RGB}{58, 107, 165}   
\definecolor{paperred}{RGB}{180, 82, 82}     
\definecolor{papergray}{RGB}{112, 128, 144}  
\definecolor{paperbg}{RGB}{245, 247, 249}

\definecolor{fbApp}{HTML}{c8e7fa}
\definecolor{fbPurple3}{HTML}{f0ebf5}

\definecolor{citecolor}{HTML}{0071BC}
\definecolor{linkcolor}{HTML}{ED1C24}

\definecolor{citecolor}{HTML}{0071BC}
\definecolor{linkcolor}{HTML}{ED1C24}

\title{XRZero-G0: Pushing the Frontier of Dexterous Robotic Manipulation with Interfaces, Quality and Ratios}

\author[]{James Wang}
\author[]{Primo Pu}
\author[]{Zephyr Fung}
\author[]{Alex Wang}
\author[]{Sam Wang}
\author[]{Bender Deng}
\author[]{Kevin Wang}
\author[]{Zivid Liu}
\author[]{Chris Pan}
\author[]{Panda Yang}
\author[]{Andy Zhai}
\author[]{Lucy Liang}
\author[]{Shalfun Li}
\author[]{Johnny Sun}
\author[]{Jacky Xu}
\author[]{Will Tian}
\author[]{Kai Yan}
\author[]{Kohler Ye}
\author[]{Scott Li}
\author[]{Qian Wang}
\author[\dagger]{Roy Gan}
\author[\ddagger]{Hao Wang}

\affiliation[]{\textbf{X SQUARE ROBOT}}
\contribution[\dagger]{Project Lead}
\contribution[\ddagger]{Correspondence}

\abstract{
The acquisition of high-quality, action-aligned demonstration data remains a fundamental bottleneck in scaling foundation models for dexterous robot manipulation. Although robot-free human demonstrations (e.g., the UMI paradigm) offer a scalable alternative to traditional teleoperation, current systems are constrained by sub-optimal hardware ergonomics, open-loop workflows, and a lack of systematic data-mixing strategies. To address these limitations, we present XRZero-G0, a hardware-software co-designed system for embodied data collection and policy learning. The system features an ergonomic, virtual reality interface equipped with a top-view camera and dual specialized grippers to directly improve collection efficiency. To ensure dataset reliability, we propose a closed-loop collection, inspection, training, and evaluation pipeline for non-proprioceptive data. This workflow achieves an 85\% data validity rate and establishes a transparent mechanism for quality control. Furthermore, we investigate the empirical scaling behaviors and optimal mixing ratios of robot-free data. Extensive experiments indicate that combining a minimal volume of real-robot data with large-scale robot-free data (e.g., a 10:1 ratio)  achieves performance comparable to exclusively real-robot datasets, while reducing acquisition costs by a factor of twenty. Utilizing XRZero-G0, we construct a 2,000-hour robot-free dataset that enables zero-shot cross-embodiment transfer to a target physical robot, demonstrating a highly scalable methodology for generalized real-world manipulation.

}
  
\date{April 15, 2026}

\metadata[Project Repo]{\url{https://github.com/X-Square-Robot/XRZero-G0}}

\begin{document}
\maketitle

\begin{figure*}[htbp]
\centering
\includegraphics[width=\textwidth]{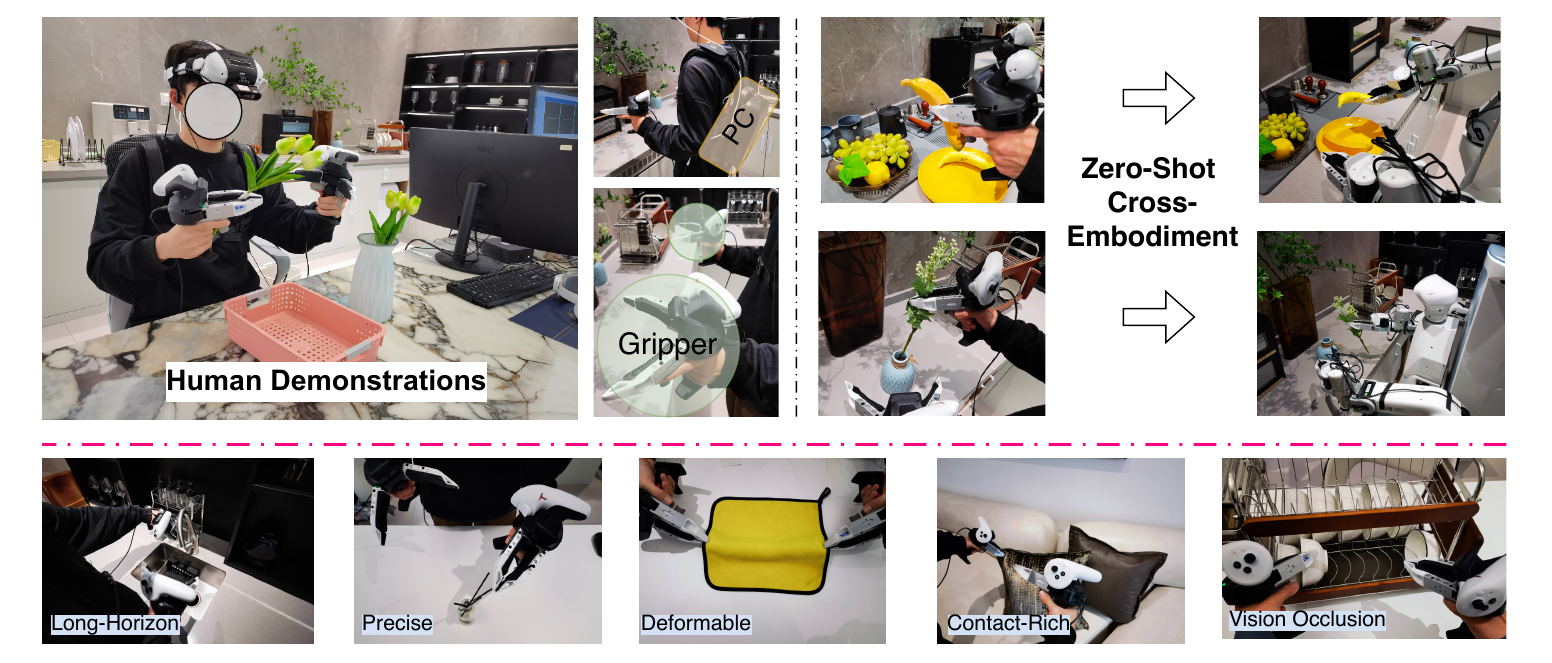}
\caption{\textbf{XRZero-G0 enables scalable robot-free data collection and cross-embodiment policy transfer.} An operator utilizes an untethered wearable system comprising a VR headset for robust spatial tracking, versatile manual grippers, and a backpack computing unit. Trajectories captured by this system map directly onto a physical dual-arm robot for policy training, facilitating zero-shot cross-embodiment execution. The system effectively captures high-fidelity demonstrations across diverse conditions, including long-horizon, precision-demanding, deformable, contact-rich, and visually occluded tasks.}
\label{fig:head}
\end{figure*}

\section{Introduction}
The rapid development of generalist robot foundation models \cite{gr00tn1_2025,wang2026latentvla,saycan,3dvla,kim2026cosmos} relies intrinsically on the availability of massive, high-fidelity human demonstration datasets \cite{openxembodiment2023}. For complex dexterous manipulation, scaling data collection while preserving precise human-robot kinematic alignment remains a formidable challenge. Traditional paradigms, such as master-slave teleoperation or VR teleoperation, provide accurate proprioceptive data but are constrained by spatial limitations, high hardware costs, and low collection throughput. Consequently, portable, robot-free data collection interfaces (e.g., the UMI paradigm \cite{ha2024umi,liu2024fastumi}) have emerged as a highly scalable alternative. By utilizing sensorized handheld devices, this paradigm decouples data acquisition from robotic hardware limits and unlocks the potential for large-scale, in-the-wild manipulation capture.

Despite their significant potential, current robot-free frameworks exhibit critical bottlenecks across three primary dimensions. First, regarding \textit{Interfaces}, existing handheld systems typically rely on visual SLAM for pose estimation. This approach is highly susceptible to tracking drift in visually degraded or dynamic environments, often restricting continuous, long-horizon data collection and causing operational fatigue. Second, concerning \textit{Data Quality}, current data processing pipelines frequently operate in an open-loop manner. The absence of comprehensive, automated quality inspection creates a reliability bottleneck where subtle kinematic anomalies remain undetected, ultimately degrading policy convergence. Finally, regarding \textit{Ratios}, the community lacks systematic empirical guidelines on how to optimally combine cost-effective robot-free demonstrations with precise, yet expensive, real-robot data. Without principled scaling behaviors, the utility of non-proprioceptive data is significantly constrained.
\\

\begin{mdframed}[
backgroundcolor=gray!15,
linewidth=0pt,
roundcorner=8pt,
innertopmargin=10pt,
innerbottommargin=10pt,
innerleftmargin=12pt,
innerrightmargin=12pt
]
\textit{This raises fundamental questions: \textcolor{paperblue}{what constitutes an optimal hardware interface for robust and continuous collection}, \textcolor{paperred}{how can we systematically guarantee dataset quality}, and \textcolor{papergray}{what data-mixing ratios best leverage non-proprioceptive demonstrations?}}
\end{mdframed}

\begin{figure}[t]
  \centering
  \includegraphics[width=\linewidth]{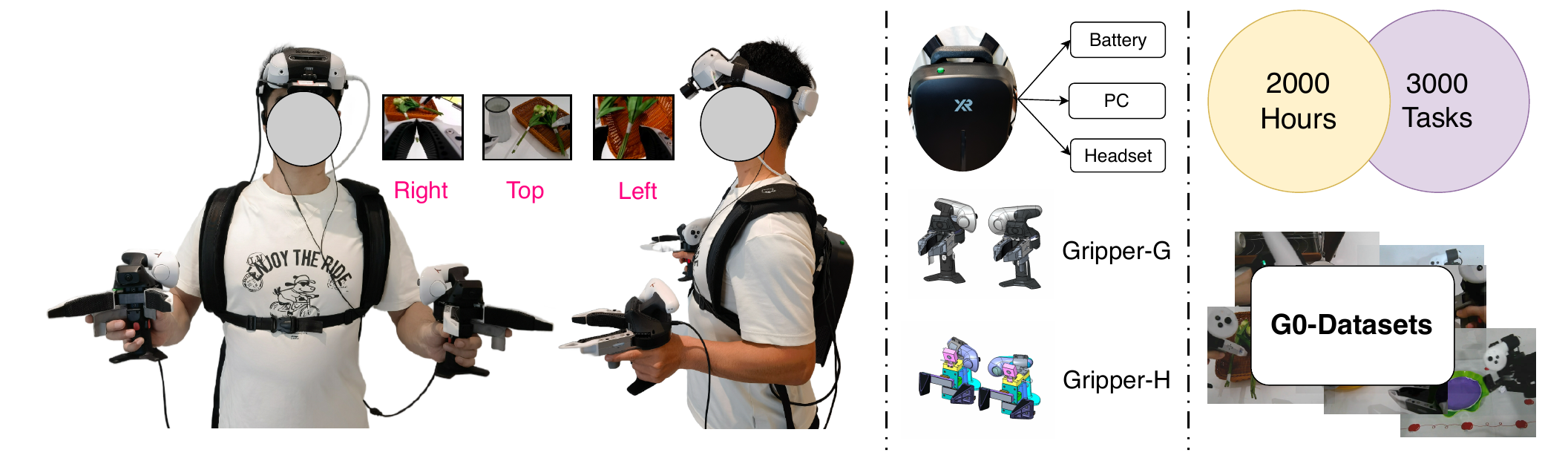}
 \caption{\textbf{The XRZero-G0 Hardware and Dataset.} An  backpack-powered VR data collection rig equipped with multi-view egocentric cameras and customized physical grippers (H \& G). This ergonomic framework enables the rapid acquisition of the large-scale G0-Dataset, encompassing 2,000 hours of multi-modal data, and 3,000 diverse manipulation tasks.}
  \label{fig:device}
\end{figure}

Our key insight is that addressing these challenges requires a tight co-design of hardware components, verification pipelines, and training methodologies. To this end, we introduce \textbf{XRZero-G0} (see Fig.~\ref{fig:head}), a comprehensive framework structured to advance dexterous manipulation through targeted innovations in interfaces, quality, and ratios. At the \textbf{\textit{Interface}} level, we pivot from unconstrained visual SLAM to a highly robust wearable architecture. By integrating a commercial VR headset for stable inside-out spatial tracking, specialized dual-mode grippers (a press-actuated H-shape and a finger-driven G-shape) for kinematic alignment (see Fig.~\ref{fig:device}), and a backpack computing unit, XRZero-G0 ensures uninterrupted, high-frequency data acquisition regardless of environmental constraints.

To guarantee superior dataset \textbf{\textit{Quality}}, we replace traditional open-loop collection with a closed-loop ``Collection-Inspection-Training-Evaluation'' pipeline. This architecture incorporates automated filtering and annotation modules to systematically identify and discard suboptimal demonstrations before they impact policy learning, significantly increasing the yield of usable training data. Furthermore, to rigorously validate the scalability of this data, we conduct systematic investigations into data \textbf{\textit{Ratios}}. We evaluate four distinct training regimes: a baseline trained exclusively on real-robot data, a zero-shot model utilizing solely robot-free data, a co-training approach for representation alignment, and an aggressive data-scaling strategy \cite{kaplan2020scaling} combining a minimal real-robot anchor with a massive volume of robot-free trajectories. Concluding this architecture is an automated benchmarking framework that delivers rapid evaluation of policy performance, completing a highly efficient iteration cycle.

The core contributions of this work are summarized as follows:
\begin{itemize}
\item \textbf{Decoupled Hardware Interfaces:} We develop a wearable hardware architecture that leverages stable VR spatial tracking and heterogeneous grippers to decouple human mobility from robotic kinematics, enabling sustained and high-throughput data capture without structural constraints.

\item \textbf{Closed-Loop Quality Verification:} We introduce a comprehensive inspection pipeline incorporating visual cleansing and inverse kinematics (IK) validation, systematically resolving the quality assurance challenges of human-centric data to yield an 85\% validity rate.

\item \textbf{Empirical Data-Mixing Laws:} We establish optimal cross-domain data-mixing guidelines, demonstrating the \textit{Few-Shot Physical Anchoring} effect. We show that augmenting a massive volume of robot-free data with a minimal real-robot anchor achieves performance comparable to purely real-robot datasets at a fraction of the cost.

\item \textbf{Large-Scale Dataset Deployment:} Leveraging this framework, we construct the G0-Dataset encompassing over 2,000 hours of validated multi-modal data. This extensive dataset supports robust spatial generalization and highly efficient cross-embodiment transfer to physical hardware, demonstrating a highly scalable methodology for real-world manipulation.
\end{itemize}
\section{Related Work}

\begin{table*}[t]
\centering
\small
\caption{Comparison of XRZero-G0 with Other UMI-based Frameworks. XRZero-G0 demonstrates superior performance in various aspects, including the highest positional accuracy (\(\leq\)4mm), support for three or more views, and the ability to natively handle multiple modalities (visual, tactile, and auditory). These unique characteristics enable XRZero-G0 to achieve high collaboration efficiency, portability, and strong cross-embodiment generalization compared to other UMI-based frameworks.}
\label{tab:comparison}
\resizebox{\textwidth}{!}{
\begin{tabular}{lcccccc}
\toprule
\textbf{Method} & \textbf{Pos. Accuracy} & \textbf{\# Views} & \textbf{Modalities (V/T/A)} & \textbf{Coll. Efficiency} & \textbf{Portability} & \textbf{Cross-Embod.} \\
\midrule
UMI \cite{chi2024universal} & \(\sim\)10\,mm & 2 & \(\checkmark\) / \(\times\) / \(\times\) & \ding{72} & \ding{72}\ding{72} & \ding{72} \\
MV-UMI \cite{rayyan2025mv} & - & 3 & \(\checkmark\) / \(\times\) / \(\times\) & \ding{72} & \ding{72}\ding{72} & \ding{72} \\
FastUMI \cite{liu2024fastumi} & \(\sim\)8\,mm & 2 & \(\checkmark\) / \(\times\) / \(\times\) & \ding{72} & \ding{72}\ding{72} & \ding{72}\ding{72} \\
ActiveUMI \cite{zeng2025activeumi} & \(\sim\)4\,mm & 3 & \(\checkmark\) / \(\times\) / \(\times\) & \ding{72}\ding{72} & \ding{72}\ding{72} & \ding{72}\ding{72} \\
DexUMI \cite{xu2025dexumi} & \(\sim\)5\,mm & 1 & \(\checkmark\) / \(\checkmark\) / \(\times\) & \ding{72}\ding{72} & \ding{72}\ding{72} & \ding{72}\ding{72} \\
UMI-FT \cite{choi2026wild}  & \(\sim\)5\,mm & 2 & \(\checkmark\) / \(\checkmark\) / \(\times\) & \ding{72}\ding{72} & \ding{72}\ding{72} & \ding{72}\ding{72} \\
TacUMI \cite{cheng2026tacumi} & \(\sim\)5\,mm & 2 & \(\checkmark\) / \(\checkmark\) / \(\times\) & \ding{72}\ding{72} & \ding{72}\ding{72} & \ding{72}\ding{72} \\
UMI-Underwater \cite{li2026umi} & - & 2 & \(\checkmark\) / \(\times\) / \(\times\) & \ding{72}\ding{72} & \ding{72}\ding{72} & \ding{72}\ding{72} \\
UMI-on-Air \cite{gupta2025umi} & - & 2 & \(\checkmark\) / \(\times\) / \(\times\) & \ding{72}\ding{72} & \ding{72}\ding{72} & \ding{72}\ding{72} \\
UMI-on-Legs \cite{ha2024umi} & - & 2 & \(\checkmark\) / \(\times\) / \(\times\) & \ding{72}\ding{72} & \ding{72}\ding{72} & \ding{72}\ding{72} \\
RDT2 \cite{liu2026rdt2} & - & 2 & \(\checkmark\) / \(\times\) / \(\times\) & \ding{72}\ding{72} & \ding{72}\ding{72} & \ding{72}\ding{72} \\
\midrule
\rowcolor{blue!8} 
\textbf{XRZero-G0 (Ours)} & \textbf{\(\le\)4\,mm} & \textbf{\(\ge\)3} & \textbf{\(\checkmark\) / \(\checkmark\) / \(\checkmark\)} & \textbf{\ding{72}\ding{72}\ding{72}} & \textbf{\ding{72}\ding{72}\ding{72}} & \textbf{\ding{72}\ding{72}\ding{72}} \\
\bottomrule
\end{tabular}
}
\end{table*}

\bigskip
The Universal Manipulation Interface (UMI) establishes the foundational framework (in Table.~\ref{tab:comparison}) for scalable, robot-free demonstration collection, recently evolving to support rapid, large-scale deployment. Pioneered by Chi et al. \cite{chi2024universal}, UMI utilizes a portable handheld gripper equipped with a wrist-mounted camera, an inertial measurement unit (IMU), and robust localization to capture high-fidelity, in-the-wild human demonstrations across unstructured environments. By employing an inference-time latency-aligned policy interface and relative-trajectory actions, UMI facilitates the generation of hardware-agnostic visuomotor policies. These policies demonstrate zero-shot transfer capabilities across diverse robotic platforms, enabling dynamic, precise, bimanual, and long-horizon behaviors without the need for robot-specific fine-tuning. Building directly upon this paradigm, FastUMI \cite{liu2024fastumi} introduces a comprehensive hardware-software redesign. This decoupled architecture simplifies system integration, minimizes setup complexity, and accommodates large-scale dataset acquisition (e.g., FastUMI-100K), thereby significantly reducing operational costs while preserving robust real-world performance.

To address contact-rich and force-sensitive manipulation, subsequent iterations of the UMI framework have integrated advanced multimodal sensing to capture fine-grained tactile and compliance feedback. For instance, UMI-FT \cite{choi2026wild} augments the baseline handheld interface with compact six-axis force/torque sensors at the fingertips, empowering policies to predict both spatial trajectories and grasping dynamics (e.g., grasp force and stiffness) for the adept execution of wiping, insertion, and skewering tasks. Similarly, TacUMI \cite{cheng2026tacumi} consolidates ViTac tactile sensors, wrist-mounted force-torque sensing, and drift-free pose tracking into a unified, compact gripper. This integration facilitates synchronized multimodal data acquisition and precise task segmentation, which are critical for complex, contact-rich operations. Furthermore, exUMI \cite{xu2025exumi} significantly improves system extensibility through AR motion-capture-based proprioception, modular visuo-tactile fingertips, rotary encoders, and automated calibration. This architecture achieves 100\% data usability and supports action-aware, task-agnostic tactile representation learning via a predictive modeling framework.

Beyond proprioceptive and tactile enhancements, recent advancements within the UMI ecosystem have expanded into active perception, dexterous manipulation, and cross-embodiment integration. ActiveUMI \cite{cheng2026tacumi} leverages a VR-augmented portable kit—featuring sensorized controllers and a head-mounted display—to explicitly record operator head movements and egocentric visual attention. This active perception approach yields policies with substantially higher in-distribution (70\%) and out-of-distribution (56\%) success rates in bimanual tasks compared to passive baselines. In the domain of dexterous manipulation, DexUMI \cite{xu2025dexumi} employs wearable exoskeletons and high-fidelity visual inpainting to utilize the human hand itself as the data collection interface, effectively minimizing kinematic and appearance discrepancies to ensure seamless skill transfer to multi-fingered robotic hands. Finally, UMI-on-Legs \cite{ha2024umi} synthesizes UMI-derived demonstrations with simulation-trained whole-body controllers, successfully bridging the embodiment gap to enable mobile manipulation on quadrupedal platforms. Collectively, these progressive developments solidify UMI as a highly versatile paradigm for abundant, high-quality, robot-free data acquisition, fundamentally driving the generalization of visuomotor learning.

\begin{figure}[t]
  \centering
  \includegraphics[width=\linewidth]{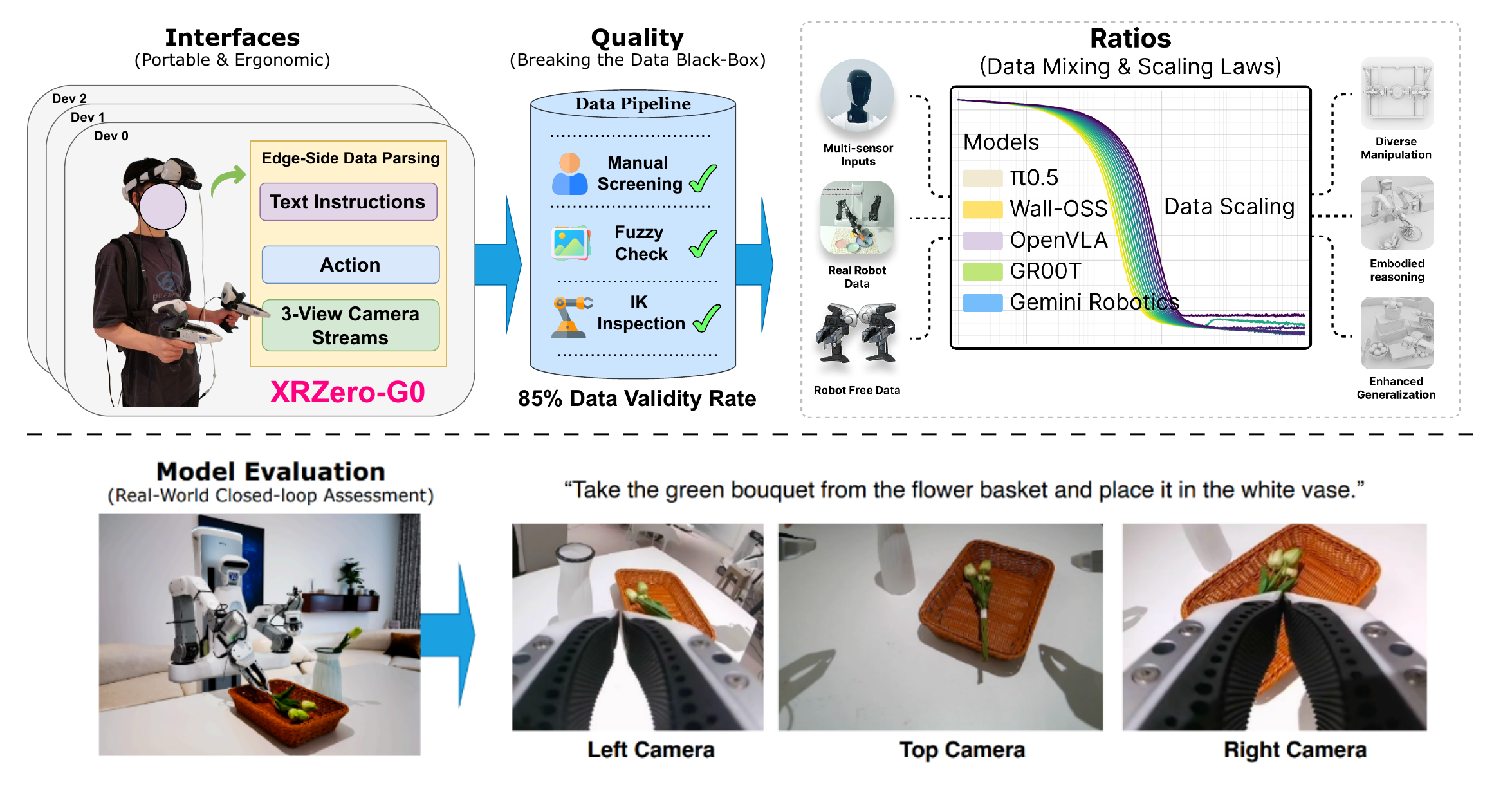}
  \caption{\textbf{Architecture of XRZero-G0.} Raw visuo-motor data is collected via an ergonomic VR interface and routed through a multi-tiered quality assurance pipeline to eliminate the data ``black-box'' problem. The resulting high-fidelity, robot-free dataset is then strategically mixed with real-robot data to train foundation policy networks, demonstrating strong positive transfer ($1+1>2$) for precise robotic manipulation.}
  \label{fig:method}
\end{figure}

\section{Method}
\label{sec:method}
This section details the architecture of \textbf{XRZero-G0}, a hardware-software co-designed framework tailored for robust data collection and policy learning (Fig.~\ref{fig:method}). To overcome the limitations of existing non-proprioceptive data paradigms, we construct a closed-loop pipeline encompassing spatial tracking, automated quality assessment, and principled data integration. The framework is systematically presented through four foundational components: portable hardware interface design, data quality verification, cross-domain data scaling strategies, and universal policy compatibility.

\subsection{Interfaces: Robust and Ergonomic Hardware Architecture}
Traditional teleoperation is physically constrained by robotic hardware, whereas recent robot-free paradigms relying on handheld monocular visual SLAM are highly susceptible to tracking drift in textureless or dynamic environments. To resolve these spatial limitations, XRZero-G0 introduces a wearable architecture that decouples human operational dexterity from robotic kinematics while maintaining rigorous tracking stability.

\begin{itemize}
\item \textbf{Multi-View Sensing and Anti-Drift Design:} The operator is equipped with a high-precision PICO 4 VR headset, which provides a genuine egocentric view via an adjustable RGB camera. To mitigate the severe visual occlusions common in complex tasks, the main egocentric view is synchronously recorded alongside dual-wrist camera feeds. This comprehensive multi-view stream provides robust visual context regardless of operational angles.

\item \textbf{Heterogeneous Grippers and Precise Pose Tracking:} We initially evaluated the tracking accuracy across various controller mounting configurations to identify the optimal position and orientation for spatial localization. To accommodate manipulation tasks of varying granularities, the system natively supports heterogeneous end-effectors. We engineered two novel physical grippers, rigidly attaching the left and right VR controllers to them: an H-shaped press-actuated gripper optimized for rapid, macroscopic grasping, and a G-shaped finger-driven gripper tailored for dexterous, fine-grained manipulation. During data acquisition, to minimize the morphological gap between human operators and the target dual-arm robot, the spatial distance between the two VR grippers is explicitly calibrated to match the baseline distance of the real robotic arms, thereby ensuring structural consistency. Concurrently, leveraging its highly stable inside-out tracking technology, the PICO 4 system delivers millimeter-accurate 6-Degree-of-Freedom (6-DoF) pose estimations in a unified world coordinate system. This facilitates the precise extraction of translational $(x, y, z)$ and rotational $(\text{roll}, \text{pitch}, \text{yaw})$ trajectories.

\item \textbf{Edge-Side Spatiotemporal Parsing:} A wearable edge computing unit manages hardware-level synchronization. This module utilizes rigorous spatiotemporal matching to precisely align natural language instructions, high-frequency 6-DoF controller trajectories, and 30Hz multi-view video streams. The synchronized data packets are subsequently transmitted to a centralized server, ensuring data fidelity while fully liberating the operator's physical workspace.
\end{itemize}

\subsection{Quality: Automated Pipeline for Data Verification}
Directly injecting raw, unverified robot-free trajectories into Vision-Language-Action (VLA) architectures introduces imperceptible noise that degrades policy convergence. XRZero-G0 mitigates this by deploying a multi-tiered post-processing pipeline designed to filter anomalies and enforce physical realism.

\begin{enumerate}
\item \textbf{Visual Cleansing and Motion Filtering:} Human kinematic frequencies frequently exceed standard robotic control limits, inducing severe motion blur. We implement automated image quality assessment algorithms to identify and discard severely blurred frames. Concurrently, stationary frames where positional variance falls below a predefined threshold are downsampled, preventing the model from internalizing passive behaviors.

\item \textbf{Kinematic Retargeting and IK Validation:} Utilizing the spatiotemporal alignment parameters and the Unified Robot Description Format (URDF) of the target embodiment, human 6-DoF trajectories are mapped directly into the robotic end-effector space. An automated Inverse Kinematics (IK) solver processes these trajectories to rigorously filter out invalid segments that violate joint limits, encounter kinematic singularities, or pose self-collision risks.

\item \textbf{Physical Playback Verification:} To guarantee high-fidelity translation to physical hardware, we establish a closed-loop playback protocol. For each task category, a subset of the filtered trajectories is randomly sampled and replayed strictly open-loop on the target dual-arm robot. Successful task completion during this physical execution serves as the definitive benchmark for trajectory validity, ensuring that the remaining dataset is fundamentally executable.

\item \textbf{Semantic Annotation:} Prolonged continuous trajectories are segmented into discrete sub-task chunks. We augment these segments with fine-grained semantic annotations detailing manipulated objects and critical keyframes, facilitating both large-scale pre-training and task-specific fine-tuning.
\end{enumerate}

Through this rigorous verification pipeline, the framework achieves a data validity rate of up to 85\%, establishing a robust foundation for downstream policy learning.

\subsection{Ratios: Principled Data Mixing and Scaling Strategies}
A central challenge in scaling embodied foundation models is determining the optimal algorithmic integration of massive, low-cost robot-free data with precise, high-cost real-robot demonstrations. We formulate a synergistic data-mixing methodology where these two domains fulfill complementary roles across distinct training phases.

\begin{itemize}
\item \textbf{Generalizable Latent Space Construction via Pre-Training:} During the initial training phase, large volumes of XRZero-G0 data serve as a semantic and spatial generalization engine. Exposed to extensive environmental diversity, the foundational model acquires robust visual-semantic alignment, object affordance recognition, and topological trajectory planning capabilities. This phase establishes a generalized multi-modal representation independent of specific robotic hardware.

\item \textbf{Kinematic Grounding via Proportionate Fine-Tuning:} While robot-free data provides profound environmental cognition, it inherently lacks embodiment-specific low-level physical priors. In the subsequent fine-tuning phase, a highly constrained proportion of real-robot data is introduced as a kinematic anchor. This physically accurate subset guides the pre-trained latent space to converge and align with the specific motor delays, friction coefficients, and joint limits of the target hardware.

\end{itemize}

This structured mixing ratio methodology theoretically justifies how augmenting a minimal fraction of real-robot data with an overwhelming volume of robot-free trajectories enables rapid zero-shot adaptation to novel environments, a claim empirically validated in Section~\ref{sec:exp}.

\subsection{Policy Network: Universal Model Compatibility}

XRZero-G0 is fundamentally designed to output a universal, model-agnostic embodied dataset. Rather than being tightly coupled to a specific learning algorithm, the meticulously verified multi-modal data streams (comprising synchronized multi-view images, language instructions, and IK-validated 6-DoF trajectories) are formatted to provide standardized supervision. This ensures direct compatibility with the two predominant policy paradigms driving contemporary robotics:

\begin{itemize}
\item \textbf{Vision-Language-Action (VLA) Paradigms:} Mainstream end-to-end continuous control models require massive quantities of precise visuo-motor pairings. The XRZero-G0 dataset natively fulfills this requirement. The comprehensive spatial context from the 3-view camera configuration directly enhances robust visual feature extraction, while the rigorously filtered 6-DoF pose trajectories serve as clean, noise-free action labels. This high-fidelity pairing allows VLA architectures to map observations to accurate end-effector movements without the performance degradation typically caused by anomalous human demonstrations.

\item \textbf{World Action Model (WAM) Paradigms:} Emerging architectures incorporating predictive world models rely heavily on understanding physical transition dynamics. Our dataset naturally supports this objective by supplying stable, high-frequency visual streams alongside fine-grained semantic annotations and discrete temporal segmentations. This structured multi-modal temporal data explicitly assists WAMs in establishing rigorous causal relationships between actions and subsequent environmental state changes, thereby enabling the forward-predictive planning required for complex, long-horizon manipulation.
\end{itemize}
\section{Data Composition}
\label{sec:data}

\subsection{Dataset Statistics and Distribution}
To validate the efficacy of the XRZero-G0 framework, we constructed the comprehensive \textbf{G0-Dataset}. Utilizing the wearable rig, human operators collected over 2,000 hours of high-fidelity, multi-modal demonstration data. The dataset spans 3,000 distinct manipulation tasks, capturing extensive object interactions within complex, unstructured environments.

\begin{figure}[t]
\centering
\includegraphics[width=\linewidth]{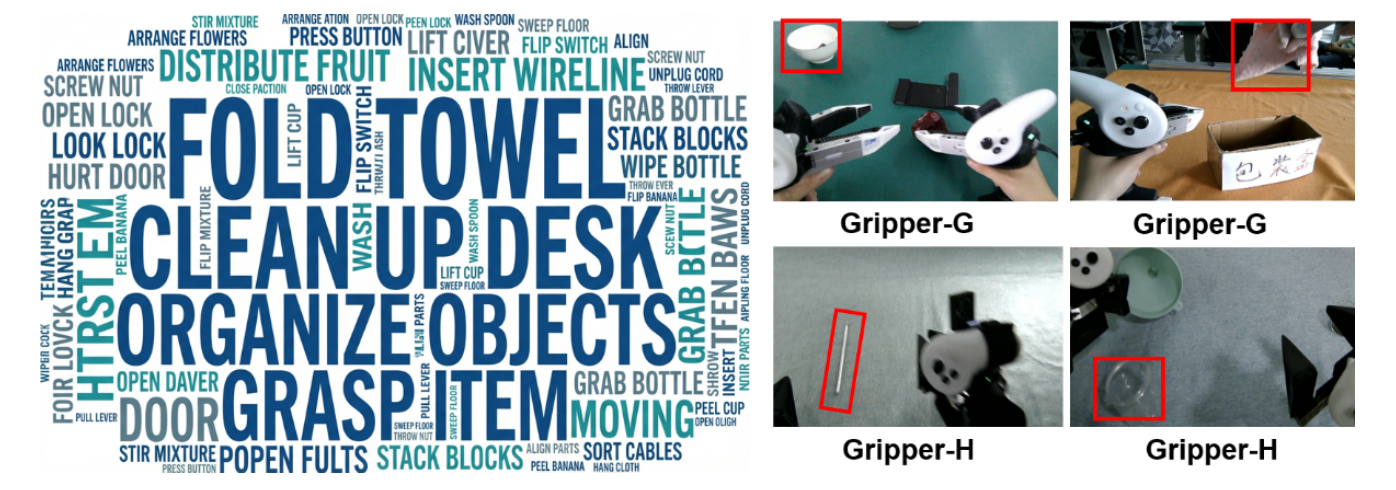}
\caption{\textbf{Overview of the G0-Dataset.} \textbf{Left:} A word cloud illustrating the pronounced long-tail task distribution, heavily featuring fundamental operations alongside thousands of specialized semantic tasks. \textbf{Right:} Egocentric visual perspectives captured during data acquisition. The frames demonstrate the deployment of the heterogeneous physical end-effectors (Gripper-G for dexterous manipulation and Gripper-H for macroscopic grasps), augmented with semantic bounding boxes highlighting the target objects for precise multi-modal supervision.}
\label{fig:data}
\end{figure}

\textbf{Task Distribution:} The dataset composition exhibits a pronounced long-tail distribution (Fig.~\ref{fig:data}). The head of the distribution is dominated by fundamental, highly repeatable tasks (e.g., fold towel,'' clean up desk,'' and ``organize objects''). These substantial quantities of common interactions enable the policy network to learn robust visual-semantic alignments and universal spatial affordances. Conversely, the long tail consists of thousands of fine-grained, specialized tasks. This structural diversity ensures extensive semantic coverage, facilitating the generalization capabilities of the model to novel operational scenarios.

\textbf{Collection Efficiency:} A primary structural advantage of this robot-free wearable paradigm is its exceptional data acquisition throughput. Based on dedicated continuous collection benchmarking across representative daily tasks, operators achieved a peak collection speed of up to 93.2 episodes per hour. This high-throughput efficiency effectively resolves the spatial footprint limitations and operational bottlenecks inherent to traditional real-robot teleoperation, demonstrating the practical viability of scaling embodied data by deeply leveraging human kinematics.

\section{Experiments}
\label{sec:exp}
In this section, we comprehensively evaluate the effectiveness and universality of the proposed XRZero-G0 framework through real-world physical experiments. The experimental design is formulated to systematically answer the following four core research questions (RQs):

\begin{itemize}
\item \textbf{RQ1 (Collection Efficiency):} Compared to traditional physical master-slave and standard VR-based teleoperation paradigms, what are the quantitative and qualitative advantages of the XRZero-G0 system regarding usability and data acquisition throughput?
\item \textbf{RQ2 (Cross-Embodiment Fidelity):} Can the human-centric data collected by this framework enable high-fidelity trajectory replay and direct policy inference across heterogeneous robotic embodiments?
\item \textbf{RQ3 (Pure Robot-Free Data Scaling):} Is it feasible to execute complex long-horizon tasks relying exclusively on pure robot-free data? Furthermore, does scaling the volume of this pure data yield a stable performance improvement?
\item \textbf{RQ4 (Data Mixing Laws):} What is the optimal quantitative mixing strategy for combining human-centric robot-free data with embodiment-specific teleoperation data to maximize policy robustness?
\end{itemize}

\subsection{Experimental Setup}
\label{subsec:exp_setup}

\textbf{Embodiments:} To rigorously validate cross-embodiment generalization, we deploy two structurally distinct dual-arm robotic systems. The CX001 platform emphasizes high dexterity with a multi-joint configuration, while the EX001 platform is designed for heavy payloads and an extended operational workspace. Both embodiments utilize the heterogeneous physical grippers detailed in Section~\ref{sec:data}.

\textbf{Base Policy Networks:} We evaluate the data pipeline using three contemporary vision-language-action (VLA) foundation models to demonstrate broad architectural compatibility:

\begin{itemize}
\item \textbf{Wall-OSS \cite{zhai2025igniting}:} An end-to-end embodied foundation model incorporating a Unified Cross-layer Chain-of-Thought (Uni-CoT) mechanism, designed for robust language-action alignment and complex 3D spatial reasoning.
\item \textbf{$\pi_0$ \cite{pi0}:} A general-purpose robot foundation model utilizing a flow-matching architecture to generate low-level motor commands from visual observations.
\item \textbf{$\pi_{0.5}$ \cite{pi05}:} An advanced iteration of $\pi_0$ co-trained on heterogeneous multi-robot datasets and web-scale data, optimized for open-world generalization.
\end{itemize}

\begin{figure}[htbp]
\centering
\includegraphics[width=\linewidth]{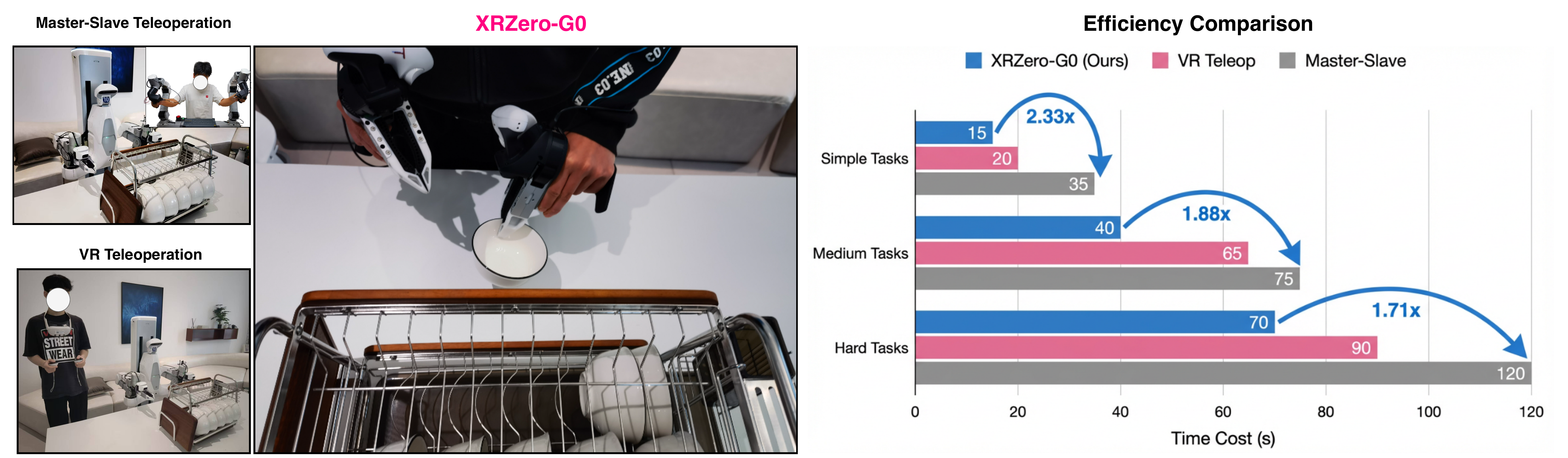} 
\caption{\textbf{Quantitative Usability and Efficiency Comparison (RQ1).} The XRZero-G0 framework demonstrates significant temporal advantages over traditional Master-Slave and standard VR teleoperation paradigms. By ergonomically decoupling human mobility from rigid robotic kinematics, the system achieves remarkable collection speedups across varying task complexities (up to 2.33$\times$ in simple tasks), validating its high-throughput viability.}
\label{fig:efficiency}
\end{figure}

\subsection{RQ1: Usability and Data Collection Efficiency}
\label{subsec:rq1_efficiency}
To address RQ1, we conduct a comparative temporal analysis evaluating the XRZero-G0 system against traditional physical master-slave teleoperation and standard VR teleoperation paradigms. Figure~\ref{fig:efficiency} illustrates the average time required to complete manipulation tasks across varying difficulty levels.

\textbf{Quantitative Efficiency:} Empirical results demonstrate that the XRZero-G0 framework yields substantial improvements in collection throughput. Specifically, compared to the traditional master-slave setup, the proposed system reduces the average completion time from 35s to 15s for simple tasks, from 75s to 40s for medium tasks, and from 120s to 70s for hard tasks. This translates to relative efficiency speedups of 2.33$\times$, 1.88$\times$, and 1.71$\times$, respectively.

\textbf{Qualitative Ergonomics:} The observed efficiency gains are primarily attributed to the ergonomic decoupling of human mobility from robotic kinematics. Traditional master-slave systems force operators to map human arm trajectories to structurally disparate robot joints, inducing high cognitive load. While standard VR teleoperation mitigates spatial constraints, it relies on virtual controllers that deprive the operator of essential tactile feedback. In contrast, XRZero-G0 employs a backpack-powered rig integrated with customized physical grippers. This configuration preserves authentic proprioceptive feedback and unconstrained first-person mobility, effectively minimizing the translation gap between human cognitive intention and physical execution.

\begin{figure}[htbp]
\centering
\includegraphics[width=\linewidth]{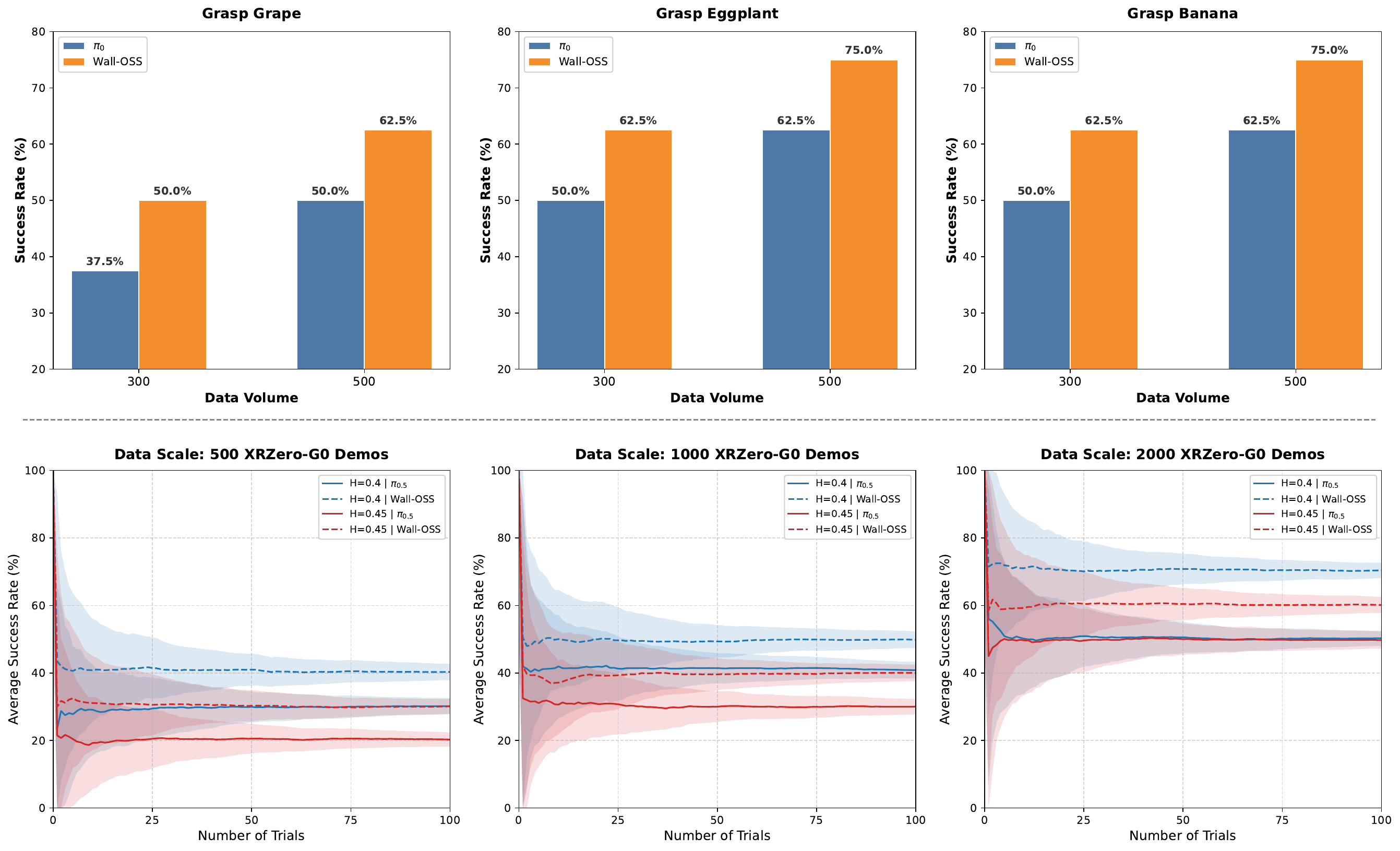} 
\caption{\textbf{Scaling Laws in Pure Robot-Free Regimes (RQ2 \& RQ3).} \textbf{Top:} Foundational grasping tasks exhibit a positive linear scaling correlation utilizing exclusively pure robot-free data (up to 500 episodes). \textbf{Bottom:} Evaluation of a complex long-horizon dual-arm task. Scaling the pure robot-free dataset up to 2,000 episodes yields stable convergence. Notably, Wall-OSS demonstrates robust 3D spatial generalization across varying operation heights ($H=0.4$m and $0.45$m), effectively overcoming the fixed-base spatial overfitting typical of traditional teleoperation datasets.}
\label{fig:pure_data_evaluation}
\end{figure}

\subsection{RQ2 \& RQ3: High-Fidelity Replay and Pure Robot-Free Policy Inference}
\label{subsec:rq2_rq3_pure_data}

To answer RQ2, we first validate the spatial transferability of the XRZero-G0 data. Subsequently, for RQ3, we investigate the feasibility of training VLA models exclusively on pure robot-free data, sequentially evaluating foundational grasping and complex dual-arm manipulation.

\textbf{Strict Data Equivalence and Trajectory Replay (RQ2):} A primary concern in human-centric data collection is the physical realizability of the generated trajectories on actual robots. The XRZero-G0 system precisely captures 6-DoF spatial poses within a strictly calibrated world coordinate frame. Through inverse kinematics (IK), these trajectories map directly to the end-effectors of heterogeneous platforms (CX001 and EX001). Real-world validation confirms that the collected data supports precise one-to-one (1:1) spatial replay, proving that the robot-free data is functionally equivalent to conventional real-robot teleoperation data.

\textbf{Scaling Laws in Foundational Tasks (RQ3):} Building upon this equivalence, we conduct policy inference experiments utilizing exclusively pure robot-free data. Models are evaluated on three geometrically diverse grasping tasks: Grasp Grape, Grasp Eggplant, and Grasp Banana. As depicted in the top row of Figure~\ref{fig:pure_data_evaluation}, scaling the pure robot-free data volume from 300 to 500 episodes yields a positive linear improvement in success rates for both $\pi_0$ and Wall-OSS. Notably, Wall-OSS achieves a 75.0\% success rate at 500 episodes in the Eggplant and Banana tasks, demonstrating superior 3D spatial alignment.

\textbf{Scaling to Complex Long-Horizon Tasks (RQ3):} To push the limits of pure robot-free data, we evaluate a demanding dual-arm Flower Arrangement task. Because human operators wear the XRZero-G0 backpack, their unconstrained bodily movements introduce dynamic variations in camera perspective and operational height. This inherently breaks the fixed-base assumption prevalent in traditional robotic datasets.

To test spatial invariance, we scale the pure robot-free dataset significantly (up to 2,000 pure robot-free episodes) and evaluate the models at two distinct robot heights ($H=0.4$m and $H=0.45$m). The running average curves (bottom row of Figure~\ref{fig:pure_data_evaluation}) reveal that Wall-OSS exhibits robust convergence as data scales. At 2,000 pure robot-free episodes, Wall-OSS achieves a 70\% success rate at $H=0.4$m and maintains a 60\% success rate at the unseen $H=0.45$m height. This confirms that sufficient volumes of unconstrained, pure robot-free data inherently endow the policy with true 3D spatial robustness, effectively overcoming the brittle spatial overfitting common in fixed-base teleoperation.

\begin{figure*}[!t] 
  \centering
  
  \includegraphics[width=0.7\linewidth]{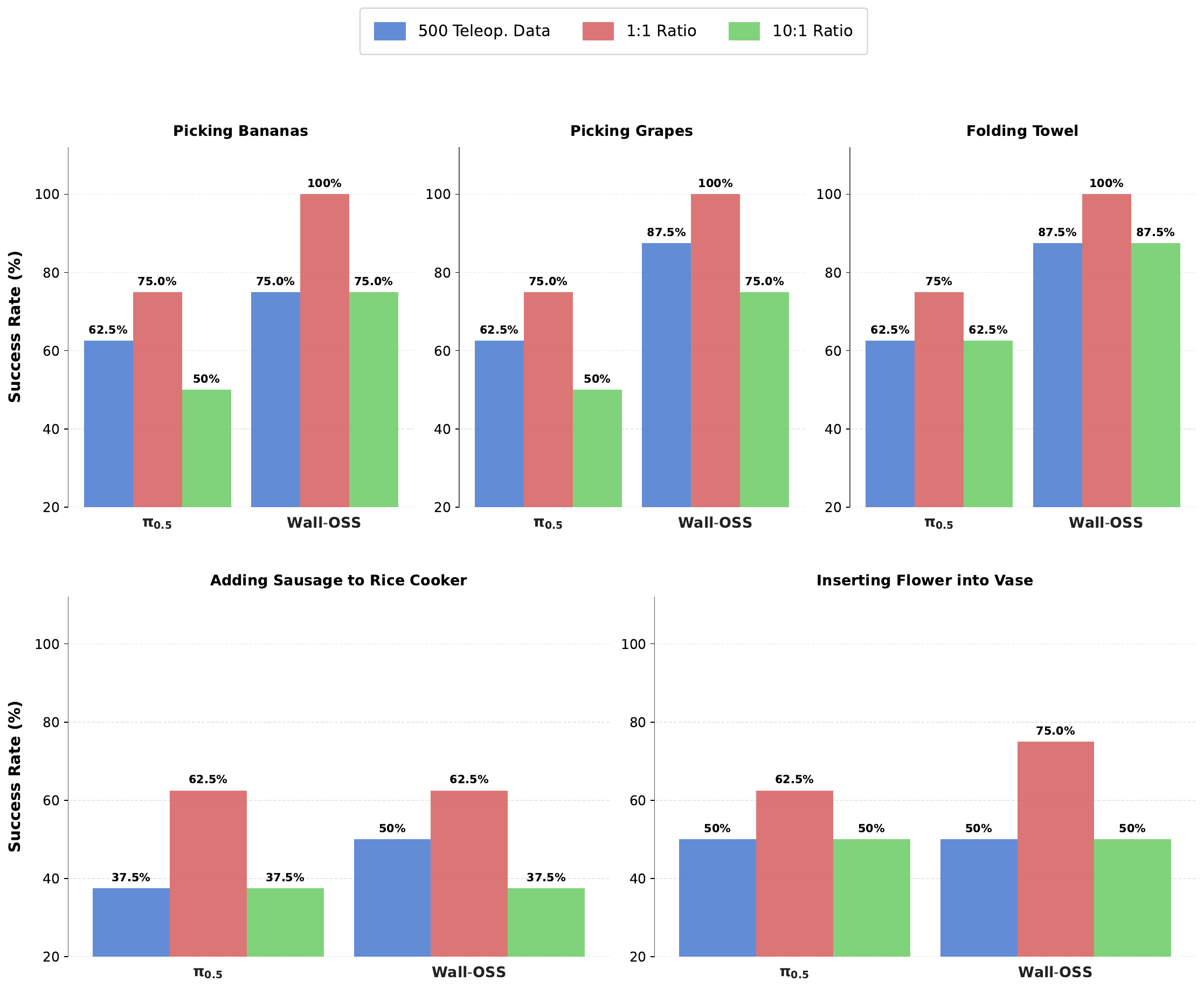} 
  \vspace{-0.2cm} 
  \caption{\textbf{Empirical Analysis of Data Mixing Strategies and Cost-Efficiency (RQ4).} Compared to a 500 real-robot baseline, Data Augmentation (1:1) adds 500 robot-free episodes to raise the performance ceiling. Meanwhile, Cost-Substitution (10:1) (500 robot-free + 50 real-robot) maintains a comparable total dataset volume while matching baseline performance. This demonstrates that few-shot physical anchoring effectively replaces 90\% of expensive real-robot data without capability loss.}
  \label{fig:data_mixing}
  
  \vspace{0.6cm} 
  
  \includegraphics[width=0.7\linewidth]{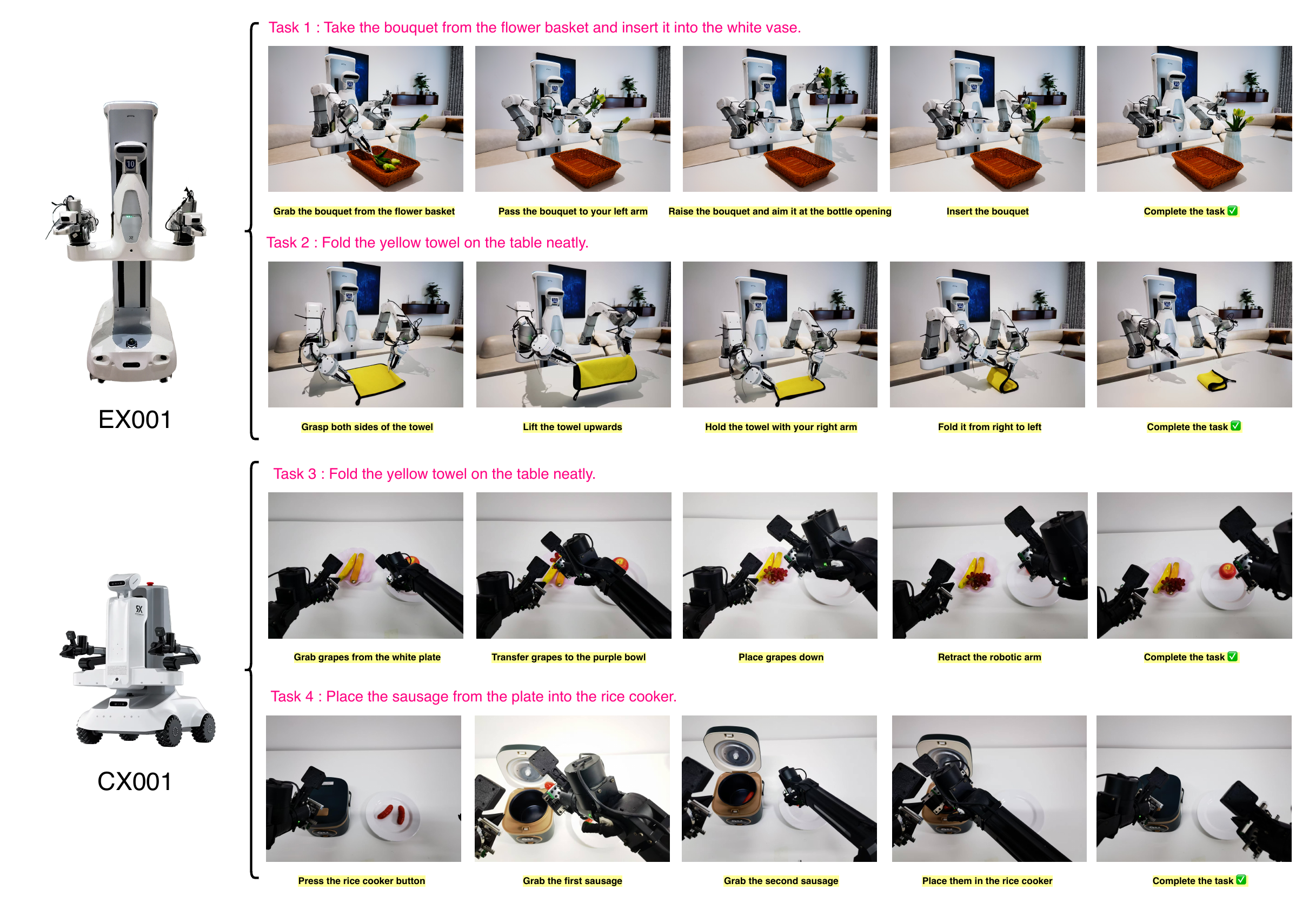} 
  \vspace{-0.2cm}
  \caption{\textbf{Zero-Shot Cross-Embodiment Execution.} Qualitative sequences of physical rollouts utilizing policies trained under the optimal 1:1 mixing strategy. The framework successfully translates multi-modal spatial intentions to structurally heterogeneous dual-arm platforms (EX001 and CX001) for complex tasks, including long-horizon coordination (Task 1), highly deformable manipulation (Tasks 2 and 3), and sequential tool interaction (Task 4).}
  \label{fig:rollouts}
  
\end{figure*}

\subsection{RQ4: Data Mixing Laws and Cost-Efficiency Paradigms}
\label{subsec:rq4_mixing}

To address RQ4, we systematically investigate the optimal scaling strategies for amalgamating cost-effective robot-free data with high-fidelity, expensive real-robot teleoperation data. Inspired by the morphological gap highlighted in recent human-centric collection systems (e.g., UMI), we hypothesize that while robot-free data provides immense semantic diversity and spatial affordances, real-robot teleoperation data remains indispensable as a kinematic anchor to provide target-specific physical priors (e.g., joint friction, PID controller delays, and kinematic singularities).

To rigorously quantify this interaction without introducing confounding variables related to total dataset size, we define a pure real-robot baseline comprising exactly 500 teleoperation episodes (\textbf{500 Teleop. Baseline}). We evaluate our data-mixing strategies against this baseline through two distinct experimental paradigms:

\begin{itemize}
    \item \textbf{Data Augmentation Paradigm (1:1 Ratio):} Consisting of the 500 baseline real-robot episodes augmented with an additional 500 robot-free episodes (Total Volume: 1,000). The objective here is to determine whether cheap robot-free data can act as a cognitive amplifier to push the performance ceiling of an existing, fixed-size real-robot dataset.
    \item \textbf{Cost-Substitution Paradigm (10:1 Ratio):} Consisting of 500 robot-free episodes explicitly anchored by a minimal footprint of only 50 real-robot episodes (Total Volume: 550). Since the total dataset volume (550) is strictly comparable to the baseline (500), this paradigm rigorously isolates the efficacy of the \textit{Few-Shot Anchoring} mechanism, investigating whether cheap data can replace 90\% of the expensive kinematic data without performance degradation.
\end{itemize}

Models are evaluated across five diverse manipulation tasks (Figure~\ref{fig:data_mixing}).

\textbf{The Augmentation Ceiling (1:1 Ratio):}
Across all evaluated tasks, augmenting the fixed 500 robotic anchor trajectories with an equivalent volume of 500 robot-free trajectories triggers a substantial performance enhancement. In the precision-demanding \textit{Inserting Flower into Vase} task, the Wall-OSS success rate increases from 50\% (pure teleoperation) to 75\% (1:1 mixed regime). This corroborates that even when real-robot data is abundant, robot-free data continues to function as a powerful cognitive amplifier, supplying robust visual-semantic representations that maximize the utility of physical demonstrations without diluting their kinematic accuracy.

\textbf{Zero-Degradation Cost-Substitution (10:1 Ratio):}
The most profound insight emerges from the 10:1 cost-substitution regime. In this setup, the absolute volume of expensive real-robot data is drastically reduced to merely 50 episodes (a 90\% reduction), while maintaining a comparable total dataset size (550 vs. 500 baseline). Remarkably, the 10:1 policy demonstrates an exceptional capability to \textbf{match or closely approach the pure 500 real-robot baseline}. For instance, in the \textit{Folding Towel} task, Wall-OSS achieves an 87.5\% success rate under the 10:1 regime, identical to the performance yielded by 500 pure real-robot episodes. Similarly, in the \textit{Picking Bananas} task, Wall-OSS achieves 75.0\% in both the pure baseline and the 10:1 mixed setup. 

We attribute this phenomenon to \textit{Few-Shot Physical Anchoring}. The 500 robot-free episodes provide sufficient environmental variance and spatial awareness, allowing the network to construct a generalized affordance manifold. Consequently, the network requires only a minimal volume of real-robot demonstrations (50 episodes) to successfully fine-tune its low-level control policies to the specific hardware kinematics.

\textbf{Economic Viability for Embodied AI:}
Considering equipment maintenance, platform development, and human operational constraints, the acquisition cost of XRZero-G0 robot-free data is approximately one-twentieth ($\frac{1}{20}$) that of traditional real-robot teleoperation. Demonstrating that a cost-substituted composition of 500 low-cost robot-free episodes and 50 high-cost real-robot episodes yields comparable efficacy to 500 pure high-cost real-robot episodes proves the tremendous economic viability of our framework. It presents a highly scalable paradigm for democratizing embodied AI data collection.

\textbf{Qualitative Cross-Embodiment Validation:}
To validate the real-world execution stability of policies trained under these mixed paradigms, Figure~\ref{fig:rollouts} illustrates continuous physical rollouts on heterogeneous platforms (EX001 and CX001). The system reliably executes intricate multi-stage behaviors, confirming that leveraging extensive robot-free data coupled with sparse physical anchoring effectively bridges the morphological gap.
\section{Conclusion and Future Work}
\label{sec:conclusion}

\subsection{Conclusion}
Scaling generalist robot foundation models for dexterous manipulation has long been bottlenecked by the prohibitive acquisition costs and ergonomic constraints associated with high-fidelity, action-aligned demonstration data. In this work, we presented \textbf{XRZero-G0}, a hardware-software co-designed framework that fundamentally addresses these bottlenecks through innovations in decoupled human-centric interfaces, closed-loop validation, and cost-efficient data amalgamation.

First, our backpack-powered VR interface equipped with multi-view egocentric perception and heterogeneous physical grippers successfully decoupled human manipulation from rigid robotic kinematics. This design not only minimized operator fatigue but also yielded a highly efficient collection throughput (93.2 episodes/hour). Second, to resolve the prevalent ``quality black box'' inherent in human-centric data, we introduced an automated Collection-Inspection-Training-Evaluation pipeline. By incorporating rigorous visual cleansing, spatial retargeting, and physical playback validation, the system maintained an 85\% deterministic spatial validity rate. 

Crucially, we systematically decoded the scaling laws and economic paradigms of cross-domain data mixing. Rather than experiencing catastrophic distribution shifts under highly skewed data ratios, our empirical results demonstrated a profound \textit{Few-Shot Physical Anchoring} phenomenon. Specifically, coupling a large volume of low-cost robot-free data with a minimal footprint of real-robot demonstrations (e.g., a 10:1 ratio comprising 500 robot-free and merely 50 real-robot episodes) achieved execution success rates comparable to a pure real-robot baseline. Given that the acquisition cost of XRZero-G0 data is approximately one-twentieth ($\frac{1}{20}$) that of traditional teleoperation, this finding establishes an exceptionally cost-effective scaling paradigm. Ultimately, by culminating in the large-scale G0-Dataset, the XRZero-G0 framework successfully enabled zero-shot cross-embodiment transfer to structurally heterogeneous dual-arm robots, providing a highly scalable and commercially viable infrastructure for generalized real-world manipulation.

\subsection{Future Work}
While XRZero-G0 establishes a robust foundation for scalable Embodied AI data generation, it opens several critical avenues for future exploration. Our subsequent research will primarily focus on the following dimensions:

\begin{itemize}
    \item \textbf{Hardware Miniaturization and Tactile Integration:} Although the current backpack-powered rig ensures unconstrained mobility, the physical weight of the integrated edge-computing unit still limits ultra-long-duration collection sessions. We intend to customise ultra-lightweight compute boards. 
    
    \item \textbf{Granular Exploration of Cost-Efficiency Boundaries:} Building upon the \textit{Few-Shot Physical Anchoring} insights discussed in Section~\ref{subsec:rq4_mixing}, we will conduct more granular scaling experiments to determine the absolute theoretical lower bound of real-robot data required for successful kinematic alignment. 
    
    \item \textbf{Expansion of Task Taxonomy to Mobile Manipulation:} To further push the boundaries of spatial invariance in VLA models, future data collection will transition from static tabletop manipulation to whole-body, unconstrained mobile environments (e.g., dynamic navigation and bimanual tool-use in active industrial or domestic settings). We will specifically target extreme contact-rich tasks and the manipulation of highly deformable objects (e.g., clothing folding and liquid pouring) to enrich the morphological diversity of the G0-Dataset.
\end{itemize}

By continuously refining the hardware-software acquisition loop and probing the limits of data mixing strategies, we envision the XRZero-G0 paradigm serving as a universal, low-cost catalyst for deploying human-level dexterity across open-world robotic ecosystems.

\clearpage
\newpage
\bibliographystyle{assets/plainnat}
\bibliography{main}

@article{chi2024universal,
  title={Universal manipulation interface: In-the-wild robot teaching without in-the-wild robots},
  author={Chi, Cheng and Xu, Zhenjia and Pan, Chuer and Cousineau, Eric and Burchfiel, Benjamin and Feng, Siyuan and Tedrake, Russ and Song, Shuran},
  journal={arXiv preprint arXiv:2402.10329},
  year={2024}
}

@article{liu2024fastumi,
  title={Fastumi: A scalable and hardware-independent universal manipulation interface with dataset},
  author={Liu, Kehui and Guan, Chuyue and Jia, Zhongjie and Wu, Ziniu and Liu, Xin and Wang, Tianyu and Liang, Shuai and Chen, Pengan and Zhang, Pingrui and Song, Haoming and others},
  journal={arXiv preprint arXiv:2409.19499},
  year={2024}
}

@article{zeng2025activeumi,
  title={Activeumi: Robotic manipulation with active perception from robot-free human demonstrations},
  author={Zeng, Qiyuan and Li, Chengmeng and John, Jude St and Zhou, Zhongyi and Wen, Junjie and Feng, Guorui and Zhu, Yichen and Xu, Yi},
  journal={arXiv preprint arXiv:2510.01607},
  year={2025}
}

@article{xu2025exumi,
  title={exumi: Extensible robot teaching system with action-aware task-agnostic tactile representation},
  author={Xu, Yue and Wei, Litao and An, Pengyu and Zhang, Qingyu and Li, Yong-Lu},
  journal={arXiv preprint arXiv:2509.14688},
  year={2025}
}

@article{cheng2026tacumi,
  title={TacUMI: A Multi-Modal Universal Manipulation Interface for Contact-Rich Tasks},
  author={Cheng, Tailai and Chen, Kejia and Chen, Lingyun and Zhang, Liding and Zhang, Yue and Ling, Yao and Hamad, Mahdi and Bing, Zhenshan and Wu, Fan and Sharma, Karan and others},
  journal={arXiv preprint arXiv:2601.14550},
  year={2026}
}

@article{xu2025dexumi,
  title={Dexumi: Using human hand as the universal manipulation interface for dexterous manipulation},
  author={Xu, Mengda and Zhang, Han and Hou, Yifan and Xu, Zhenjia and Fan, Linxi and Veloso, Manuela and Song, Shuran},
  journal={arXiv preprint arXiv:2505.21864},
  year={2025}
}

@article{choi2026wild,
  title={In-the-Wild Compliant Manipulation with UMI-FT},
  author={Choi, Hojung and Hou, Yifan and Pan, Chuer and Hong, Seongheon and Patel, Austin and Xu, Xiaomeng and Cutkosky, Mark R and Song, Shuran},
  journal={arXiv preprint arXiv:2601.09988},
  year={2026}
}

@article{ha2024umi,
  title={Umi on legs: Making manipulation policies mobile with manipulation-centric whole-body controllers},
  author={Ha, Huy and Gao, Yihuai and Fu, Zipeng and Tan, Jie and Song, Shuran},
  journal={arXiv preprint arXiv:2407.10353},
  year={2024}
}

@inproceedings{wang2026latentvla,
  title={LatentVLA: Taming Latent Space for Generalizable and Long-Horizon Bimanual Manipulation},
  author={Wang, Junming},
  booktitle={Proceedings of the AAAI Conference on Artificial Intelligence},
  volume={40},
  number={22},
  pages={18593--18601},
  year={2026}
}

@article{liu2026rdt2,
  title={RDT2: Exploring the Scaling Limit of UMI Data Towards Zero-Shot Cross-Embodiment Generalization},
  author={Liu, Songming and Li, Bangguo and Ma, Kai and Wu, Lingxuan and Tan, Hengkai and Ouyang, Xiao and Su, Hang and Zhu, Jun},
  journal={arXiv preprint arXiv:2602.03310},
  year={2026}
}

@article{rayyan2025mv,
  title={MV-UMI: A Scalable Multi-View Interface for Cross-Embodiment Learning},
  author={Rayyan, Omar and Abanes, John and Hafez, Mahmoud and Tzes, Anthony and Abu-Dakka, Fares},
  journal={arXiv preprint arXiv:2509.18757},
  year={2025}
}

@article{li2026umi,
  title={UMI-Underwater: Learning Underwater Manipulation without Underwater Teleoperation},
  author={Li, Hao and Chung, Long Yin and Goler, Jack and Zhang, Ryan and Xie, Xiaochi and Ha, Huy and Song, Shuran and Cutkosky, Mark},
  journal={arXiv preprint arXiv:2603.27012},
  year={2026}
}

@article{gupta2025umi,
  title={UMI-on-Air: Embodiment-Aware Guidance for Embodiment-Agnostic Visuomotor Policies},
  author={Gupta, Harsh and Guo, Xiaofeng and Ha, Huy and Pan, Chuer and Cao, Muqing and Lee, Dongjae and Scherer, Sebastian and Song, Shuran and Shi, Guanya},
  journal={arXiv preprint arXiv:2510.02614},
  year={2025}
}

@article{kim2026cosmos,
  title={Cosmos policy: Fine-tuning video models for visuomotor control and planning},
  author={Kim, Moo Jin and Gao, Yihuai and Lin, Tsung-Yi and Lin, Yen-Chen and Ge, Yunhao and Lam, Grace and Liang, Percy and Song, Shuran and Liu, Ming-Yu and Finn, Chelsea and others},
  journal={arXiv preprint arXiv:2601.16163},
  year={2026}
}

@article{kaplan2020scaling,
  title={Scaling laws for neural language models},
  author={Kaplan, Jared and McCandlish, Sam and Henighan, Tom and Brown, Tom B and Chess, Benjamin and Child, Rewon and Gray, Scott and Radford, Alec and Wu, Jeffrey and Amodei, Dario},
  journal={arXiv preprint arXiv:2001.08361},
  year={2020}
}

@article{pi0,
  title   = {$\\pi_{0}$: A Vision-Language-Action Flow Model for General Robot Policies},
  author  = {Black, Kevin and others},
  journal = {arXiv preprint arXiv:2410.24164},
  year    = {2024}
}

@inproceedings{3dvla,
  title     = {3D-VLA: A 3D Vision-Language-Action Generative World Model},
  author    = {Zhen, Haoyu and Qiu, Xiaowen and Chen, Peihao and Yang, Jincheng and Yan, Xin and Du, Yilun and Hong, Yining and Gan, Chuang},
  booktitle = {Proceedings of the 41st International Conference on Machine Learning (ICML)},
  series    = {Proceedings of Machine Learning Research},
  volume    = {235},
  pages     = {61229--61245},
  year      = {2024},
  month     = {July},
  publisher = {PMLR},
  url       = {https://proceedings.mlr.press/v235/zhen24a.html}
}

@article{pi05,
  title   = {$\pi_{0.5}$: a Vision-Language-Action Model with Open-World Generalization},
  author  = {Physical Intelligence and Black, Kevin and Brown, Noah and others},
  journal = {arXiv preprint arXiv:2504.16054},
  year    = {2025},
  url     = {https://arxiv.org/abs/2504.16054}
}

@article{saycan,
  title   = {Do As I Can, Not As I Say: Grounding Language in Robotic Affordances},
  author  = {Ahn, Michael and others},
  journal = {arXiv preprint arXiv:2204.01691},
  year    = {2022},
  url     = {https://arxiv.org/abs/2204.01691}
}

@inproceedings{gr00tn1_2025,
  archivePrefix = {arxiv},
  eprint     = {2503.14734},
  title      = {{GR00T} {N1}: An Open Foundation Model for Generalist Humanoid Robots},
  author     = {NVIDIA and Johan Bjorck andFernando Castañeda, Nikita Cherniadev and Xingye Da and Runyu Ding and Linxi "Jim" Fan and Yu Fang and Dieter Fox and Fengyuan Hu and Spencer Huang and Joel Jang and Zhenyu Jiang and Jan Kautz and Kaushil Kundalia and Lawrence Lao and Zhiqi Li and Zongyu Lin and Kevin Lin and Guilin Liu and Edith Llontop and Loic Magne and Ajay Mandlekar and Avnish Narayan and Soroush Nasiriany and Scott Reed and You Liang Tan and Guanzhi Wang and Zu Wang and Jing Wang and Qi Wang and Jiannan Xiang and Yuqi Xie and Yinzhen Xu and Zhenjia Xu and Seonghyeon Ye and Zhiding Yu and Ao Zhang and Hao Zhang and Yizhou Zhao and Ruijie Zheng and Yuke Zhu},
  month      = {March},
  year       = {2025},
  booktitle  = {ArXiv Preprint},
}

@article{openxembodiment2023,
  title   = {Open X-Embodiment: Robotic learning datasets and RT-X models},
  author  = {Open X-Embodiment Collaboration and Padalkar, Abhishek and Pooley, Acorn and Jain, Ajinkya and Bewley, Alex and Herzog, Alex and Irpan, Alex and Khazatsky, Alexander and Rai, Anant and Singh, Anikait and Brohan, Anthony and others},
  journal = {arXiv preprint arXiv:2310.08864},
  year    = {2023}
}

@article{zhai2025igniting,
  title={Igniting vlms toward the embodied space},
  author={Zhai, Andy and Liu, Brae and Fang, Bruno and Cai, Chalse and Ma, Ellie and Yin, Ethan and Wang, Hao and Zhou, Hugo and Wang, James and Shi, Lights and others},
  journal={arXiv preprint arXiv:2509.11766},
  year={2025}
}

\end{document}